# A Survey of Toxic Comment Classification Methods

Kehan Wang, Jiaxi Yang, Hongjun Wu
{kw454, jy477, hw434} @ cornell.edu

## Abstract

While in real life everyone behaves themselves at least to some extent, it is much more difficult to expect people to behave themselves on the internet, because there are few checks or consequences for posting something toxic to others. Yet, for people on the other side, toxic texts often lead to serious psychological consequences. Detecting such toxic texts is challenging. In this paper, we attempt to build a toxicity detector using machine learning methods including CNN, Naive Bayes model, as well as LSTM. While there has been numerous groundwork laid by others, we aim to build models that provide higher accuracy than the predecessors. We produced very high accuracy models using LSTM and CNN, and compared them to the go-to solutions in language processing, the Naive Bayes model. A word embedding approach is also applied to empower the accuracy of our models.

## Introduction

The emergence of the internet and the easy access to online forums with undetectable identities has since created a problem that people abuse the chat system and send toxic messages. Detecting such toxic texts which can contain insults, bulgar words, and threats, is a difficult task. The most traditional and still being widely used method of detecting such toxic texts is by text matching. The software compares the text with an existing dictionary and checks if the text is in the library. This method quickly got exploited because people naturally want to bypass the sensoring and invent new words that have the same meaning but are spelled differently. Another common way to bypass such restriction is by shortening the word or message, which also breaks a dictionary-based toxic text detection system. The process of growing such a dictionary soon becomes tedious as new methods of combating dictionary-based toxic text systems are continually being invented by internet users every day, and it is apparent that a better context-based detection system is needed to solve this problem. Naturally, one would look towards the state-of-the-art method in computer science to solve this problem. With the dramatic increase of computing power in consumer devices, classic machine learning methods which require far more computing power than dictionary comparison methods but yield potentially better results, become a possibility.

In this paper, we will explore the classic solution using Naive Bayes for classification, as well as more advanced methods using Convolutional Neural Networks (CNNs) and Long Short Term Memory (LSTM). Our aim is to compare the accuracy between the classic Naive Bayes method to CNN and LSTM, as well as build a version of our own CNN and LSTM under the robust word embedding approach that could increase the accuracy score dramatically.

## Background

The dataset we analyzed was published by Google Jigsaw in December 2017. Each column in the data contains column names such as `id`, `comment content`, and `toxicity` (whether the poisonous text is a comment or not, or the chance that the remark is hazardous), and categories of toxicity, that includes labels `toxic`, `severe toxic`, `insult`, `threat`, `obscene`, and `identity hate`. The majority of the comments were taken from the English Wikipedia and are written in English, with a few exceptions in Arabic, Chinese, or German. The job is framed as a multi-label classification issue because comments can be associated with numerous classes at once. Official definitions for the six classes have yet to be released by Jigsaw. but they do state that they defined a toxic comment as "a rude, disrespectful, or unreasonable comment that is likely to make you leave a discussion". In order to better understand our work, a basic understanding of the selected machine learning algorithms is needed. The algorithms we used are Naive Bayes, Convolutional Neural Networks (CNNs), and Long Short Term Memory (LSTM).

## Method

1. **Data Processing**

First and foremost, we opted to check for missing values in the downloaded data after importing the training and test data into the pandas dataframe. We discovered that there were no missing records using the "isnull" function on both the training and test data. Next, we decided to normalize the text data since comments from online forums, like Wikipedia usually contain inconsistent language, use of special characters in place of letters (e,g. @username), as well as the use of numbers to represent letters(e.g. g00d). To solve this problem, we use Regex to implement text normalization. In order to train a deep-learning model with clean text input, the data must first be transformed to a machine-readable format. In our case, we use Tokenizer class from Keras library to vectorize a text corpus by turning each text into either a sequence of integers (each integer being the index of a token in a dictionary) or into a vector where the coefficient for each token could be binary, based on word count, based on tf-idf, etc. Nevertheless, a Padding strategy with `pad_sequences` function from Keras is also used in our preprocessing stage so that we can pass vectors of consistent lengths to our models which we can avoid inconsistency when variable-length sentences are converted into variable-length sequence vectors.

2. **Word Embedding with FastText and GloVe**

To harness the potential of Transfer Learning, we decided to employ FastText and GloVe, which provide pre-trained word embedding during the deep learning models implementation. FastText is essentially an extension of the word2vec model, treating each word as composed of character n-grams. So the vector for a word is made of the sum of this character n-grams. GloVe treats each word in the corpus like an atomic entity and generates a vector for each word. In this sense, GloVe is very much like word2vec- both treat words as the smallest unit to train on. We run the experiments on both word embeddings to test which are more suitable for our toxic comment classification. In the current stage, we assume FastText has better performance as it can learn representations for misspelled, obfuscated or abbreviated words which are often present in online comments based on Bojanowski's study[1][2].

3. **Naive Bayes**

According to Bayes' theorem, conditional probability is used to reversely predict the type of certain word text, which is a little different from the other two methods. A dedicated data preprocessing for the Naive Bayes algorithm is needed. We processed raw data using TF-IDF and ngrams, which converts text to a certain feature matrix including frequency of words and appearance.

---

[1] T. Mikolov, et al. Advances in Pre-Training Distributed Word Representations.
[2] Pennington, Jeffrey & Socher, Richard & Manning, Christopher. (2014). Glove: Global Vectors for Word Representation.

$$P(A|B) \;=\; P(A) \;*\; P(B|A) \;/\; P(B)$$

In the above formula, *A* represents the possibility that one can be classified as certain categories such as `toxic`, `obscene`, `threat`, etc. *B* represents the certain feature which is generated from preprocessing. By utilizing the possibilities of certain features under some category condition, we can reverse the conditional probability.

4. **CNN**

We implemented our procedure based on the standard CNN methodology[3]. We used FastText/GloVe to create an embedding layer that matches the input words with the fixed dense vector. An indicative description of the process is illustrated in the following figure.

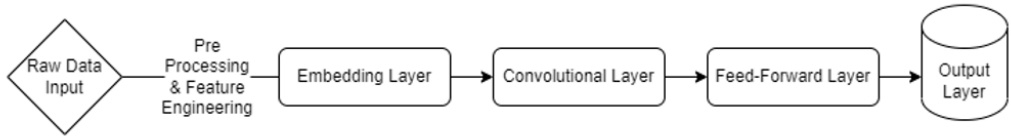

*Figure 1. CNN Structure*

The CNN model we implemented consists of one 1-dimensional convolutional layer across the concatenated word embeddings for each input comment. The convolutional layer has a fixed number of filters with a kernel size of 3. The next layer is a fully connected layer with fixed numbers units which is then followed by the output layer. We will elaborate on our hyperparameter tuning to find the best parameters that would give us the highest accuracy in the Experimental Analysis part.

5. **LSTM(Long Short Term Memory)**

The Long Short Term Memory Recurrent Neural Network is used to identify the toxic comments. The overall structure implementation of LSTM is very similar to the standard RNN. The overall structure is shown in the following figure.

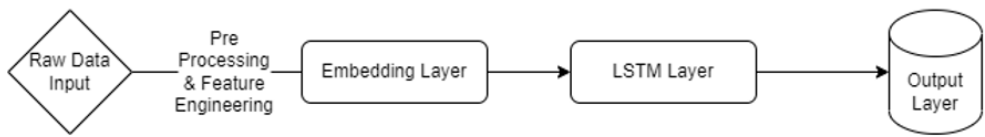

*Figure 2. LSTM Structure*

The individual cell states in this architecture have the capacity to remove or add information to the cell state through gate layers, which makes it appealing. In practice, this is important since it allows the model to recall ideas drawn from words throughout the comment. The LSTM model consists of one densely connected layer with fixed number units across the concatenated word vectors for each of the words in the comment. Again, we will use FastText/GloVe to create an embedding layer.

---

[3] Georgakopoulos, Spiros & Tasoulis, Sotiris & Vrahatis, Aristidis & Plagianakos, Vassilis.

# Experimental analysis

## 1. Experimental Setup

We mentioned our pre-processing and how we implemented the model in the Method part. During our experiments, we randomly split the data into train-set and validation-set before attempting to fit a model on the training data; the validation set accounts for 20% of the training data. In our experiment, we performed hyperparameter tuning using GridSearchCV and found the best parameters that would give me the highest accuracy. All the results shown in section 2.3 and 2.4 are calculated by the parameters that have the best performance. The sections are organized as follows: section 2.1 is about classification accuracy of five methods, section 2.2 elaborates on tuning, section 2.3 shows kaggle results of five methods, and section 2.4 concludes an overall result analysis.

## 2. Result and Analysis

### 2.1. Classification Accuracy

We first compared the model performance using the accuracy score. All five methods have exciting performances, generally over 95 percent. According to *Table 1*, which shows the average accuracy of classification for five categories of toxic. It is obvious that LSTM and CNN achieve more accuracy than Naive Bayes by 3 percent but the difference between two advanced methods is little, and models run with GloVe have slightly better performance.

| Models | Naive Bayes | LSTM(with FastText) | LSTM(with GloVe) | CNN(with FastText) | CNN(with GloVe) |
|---|---|---|---|---|---|
| Accuracy | 96.8% | 99.4% | 99.6% | 99.4% | 99.5% |

*Table 1. Prediction Accuracy Table*

### 2.2. Tuning

We applied the GridSearchCV method to find the best parameter for LSTM and CNN models, and *Figure 3* shows accuracy and performance with different hyper parameters. Looking through all these parameters we used, we can easily grab out the best hyper parameter which achieves the highest accuracy for the two models above.

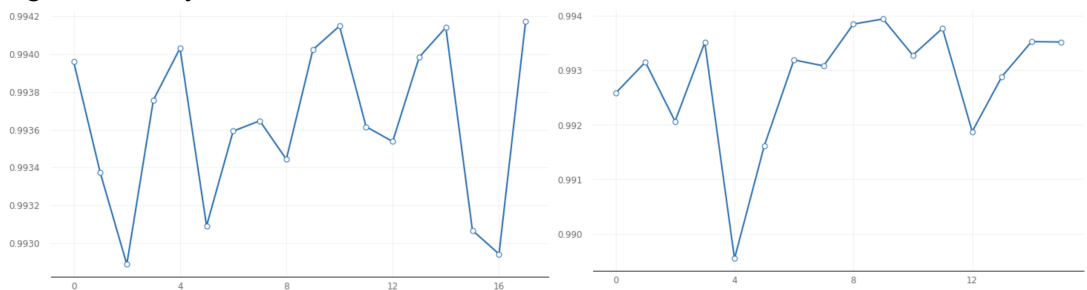

*Figure 3. Tuning Plot*

### 2.3. Kaggle Results

We also submitted our result to the Kaggle competition to check how well our models can perform. *Table 5* shows the result of our five methods in kaggle. The public score is calculated with apprimiately 10% of the test data and the private score is calculated with approximately 90% of the test data. The scores are presented on the mean column-wise ROC AUC. In other words, the score is the average of the individual AUCs of each predicted column (toxic categories).

| Model | Public Score | Private Score |
|---|---|---|
| Naive Bayes | 0.84029 | 0.83104 |
| LSTM(with FastText) | 0.94388 | 0.94584 |
| LSTM(with GloVe) | 0.97718 | 0.97665 |
| CNN(with FastText) | 0.94975 | 0.95234 |
| CNN(with GloVe) | 0.95675 | 0.95239 |

*Table 2. Kaggle Result*

### 2.4. Real-world case analysis

In order to show how our models perform clearly in the real-world case, we set up several sentences and test whether they involve certain types of toxic, with results in *Appendix A*.

# Discussion

It is more common for CNN to solve problems on image processing or recognition. But in our work, we trained a CNN with an embedding layer, which is different from prior work applying common NLP models in detecting toxicity. Our model achieves almost the same performance as traditional NLP models. While we assumed FastText would have better performance, surprisingly, our result shows models with GloVe embedding layers actually outperforms FastText. The authors suspect this performance was caused by the fact that toxic words are not considered as rare words in the FastText training process.

# Conclusion

We implemented five methods to detect toxic levels of different sentences based on online comments, and all five models achieved optimal performance. In Particular, LSTM with GloVe embedding layer achieves the best accuracy and Kaggle score, and GloVe embedding layers have an overall better performance. The authors acknowledge that there are potential areas for improvement in the data pre-processing part of the experiment, therefore one potential future work is to explore different approaches for data pre-processing. We would also attempt a very interesting approach that combined LSTM and CNN[4] which might yield additional performance gain when applied to our toxic comment classification problems.

---

[4] Jiarui Zhang, et al. LSTM-CNN Hybrid model for text classification. IEEE Xplore. (n.d.). Retrieved December 10, 2021.

# Appendix

## A. Real-World case analysis

According to the Three Tables 2,3,4, all of the advanced methods achieve quite reasonable results. For example, all the sentence with "f**k" are predicted to be toxic and obscene, and all the sentences with "kill" are predicted as threats. On the contrary, all friendly words basically have low scores on each of the types. Among these three methods, CNN achieves the best results and LSTM and Naive Bayes follow behind. It is quite surprising that Naive Bayes also gets a quite reasonable result that gives all the positive sentences with 0 percentage toxic. Therefore, by using LSTM or CNN, sentences can be detected whether they are toxic or not. And we can achieve a quite good performance by tuning the hyper parameters and more specific preprocessing. Note: All the sword words are partly replaced by '*'.

|  | go jump off a bridge j**k | i will kill you | have a nice day | f**k ofF!! | Hello, How are you? | get the f**k away from me @ss**le!! |
|---|---|---|---|---|---|---|
| **Toxic** | 55% | 72% | 2% | 97% | 1% | 53% |
| **Severe Toxic** | 0% | 0% | 0% | 1% | 0% | 0% |
| **Obscene** | 13% | 10% | 0% | 86% | 0% | 9% |
| **Threat** | 0% | 1% | 0% | 0% | 0% | 0% |
| **Insult** | 14% | 6% | 0% | 34% | 0% | 2% |
| **Identity Hate** | 0% | 0% | 0% | 0% | 0% | 0% |

*Table 3. Toxic Percentage of several sentences predicted by LSTM (with GloVe)*

|  | go jump off a bridge j**k | i will kill you | have a nice day | f**k ofF!! | Hello, How are you? | get the f**k away from me @ss**le!! |
| --- | --- | --- | --- | --- | --- | --- |
| **Toxic** | 91% | 43% | 10% | 99% | 10% | 96% |
| **Severe Toxic** | 19% | 3% | 0% | 34% | 0% | 18% |
| **Obscene** | 76% | 19% | 2% | 95% | 2% | 88% |
| **Threat** | 4% | 3% | 0% | 3% | 0% | 3% |
| **Insult** | 64% | 24% | 3% | 79% | 4% | 69% |
| **Identity Hate** | 14% | 6% | 1% | 14% | 1% | 14% |

*Table 4. Toxic Percentage of several sentences predicted by LSTM (with GloVe)*

|  | go jump off a bridge j**k | i will kill you | have a nice day | f**k ofF!! | Hello, How are you? | get the f**k away from me @ss**le!! |
| --- | --- | --- | --- | --- | --- | --- |
| **Toxic** | 100% | 83% | 1% | 100% | 8% | 100% |
| **Severe Toxic** | 34% | 10% | 0% | 37% | 0% | 20% |
| **Obscene** | 94% | 53% | 0% | 99% | 1% | 92% |
| **Threat** | 5% | 5% | 0% | 2% | 0% | 2% |
| **Insult** | 86% | 46% | 0% | 92% | 2% | 77% |
| **Identity Hate** | 18% | 11% | 0% | 12% | 1% | 9% |

*Table 5. Toxic Percentage of several sentences predicted by LSTM (with GloVe)*